\documentclass[runningheads]{llncs}
\usepackage[T1]{fontenc}
\usepackage{graphicx}
\usepackage[square,sort,comma,numbers,sectionbib]{natbib}
\usepackage{url}

\usepackage[title]{appendix}
\usepackage{xcolor}

\usepackage{booktabs}
\usepackage{multirow}

\usepackage{bm}
\usepackage{amsfonts}
\usepackage{amssymb}
\usepackage{amsmath}
\usepackage{dsfont}

\newcommand{\VX}{\boldsymbol{X}}
\newcommand{\Vx}{\boldsymbol{x}}
\newcommand{\VY}{\boldsymbol{Y}}
\newcommand{\Vy}{\boldsymbol{y}}
\newcommand{\VW}{\boldsymbol{W}}
\newcommand{\Vw}{\boldsymbol{w}}

\newcommand{\Sample}{\mathcal{O}}
\newcommand{\tHat}{\widehat{t}}
\newcommand{\VtHat}{\widehat{\boldsymbol{t}}}

\newcommand{\etaHat}{\widehat{\eta}}
\newcommand{\sigmaHat}{\widehat{\sigma}}
\newcommand{\Vsigma}{\boldsymbol{\sigma}}

\newcommand{\SigmaHat}{\widehat{\Sigma}}

\newcommand{\muHat}{\widehat{\mu}}
\newcommand{\VmuHat}{\widehat{\boldsymbol{\mu}}}

\newcommand{\psiHat}{\widehat{\psi}}
\newcommand{\VpsiHat}{\widehat{\boldsymbol{\psi}}}

\newcommand{\varphiHat}{\widehat{\varphi}}
\newcommand{\Vvarphi}{\boldsymbol{\varphi}}
\newcommand{\ThetaHat}{\widehat{\Theta}}
\newcommand{\PsiHat}{\widehat{\Psi}}

\newcommand{\Model}{\mathcal{M}}

\newcommand\Gaus{\mathcal{N}}
\newcommand\Esp{\mathbb{E}}
\newcommand\Prob{\mathbb{P}}
\newcommand\Var{\mathbb{V}ar}

\newcommand\Cov{\mathbb{C}ov}

\newcommand\Score{\mathcal{S}}

\newcommand\dpartial[2]{\frac{\partial #1}{\partial #2}}

\newcommand\trans[1]{{#1}^\intercal}

\newcommand{\Ind}{\mathds{1}}
\newcommand{\Vone}{\mathbf{1}}

\begin{document}
\title{A sensitivity analysis to quantify the impact of neuroimaging preprocessing strategies on subsequent statistical analyses}
\titlerunning{Sensitivity analysis for neuroimaging multiverse analyses}
%

\author{Brice Ozenne\inst{1,2}\orcidID{0000-0001-9694-2956} \and
Martin Nørgaard\inst{3,4}\orcidID{1111-2222-3333-4444} \and
Cyril Pernet\inst{1}\orcidID{2222--3333-4444-5555}\and
Melanie Ganz\inst{1,4}\orcidID{2222--3333-4444-5555}}

\authorrunning{B. Ozenne et al.}
%
\institute{Neurobiology Research Unit, Rigshospitalet \and
Section of Biostatistics, University of Copenhagen \and
Molecular Imaging Branch, National Institute of Mental Health (NIMH) \and
Department of Computer Science, University of Copenhagen \\
\email{brice-ozenne@nru.dk}}
\maketitle              
\begin{abstract}
Even though novel imaging techniques have been successful in studying brain structure and function, the measured biological signals are often contaminated by multiple sources of noise, arising due to e.g. head movements of the individual being scanned, limited spatial/temporal resolution, or other issues specific to each imaging technology. Data preprocessing (e.g. denoising) is therefore critical.
Preprocessing pipelines have become increasingly complex over the years, but also more flexible, and this flexibility can have a significant impact on the final results and conclusions of a given study. This large parameter space is often referred to as multiverse analyses. Here, we provide conceptual and practical tools for statistical analyses that can aggregate multiple pipeline results along with a new sensitivity analysis testing for hypotheses across pipelines such as “no effect across all pipelines” or “at least one pipeline with no effect”. The proposed framework is generic and can be applied to any multiverse scenario, but we illustrate its use based on positron emission tomography data.

\end{abstract}
\section{Introduction} 

Modern neuroimaging techniques have provided unique opportunities to measure complex signaling pathways in the living human brain with the goal of identifying reliable  biomarkers of disease states and treatment outcomes \cite{Poldrack2020}. Data arising from state-of-the-art neuroimaging techniques are, however, often contaminated with noise confounds such as motion-related artefacts, affecting both the spatial and temporal correlation structure of the data \cite{norgaard2019preprocessing}. Carefully designed preprocessing steps have been developed to remove unwanted noise sources, but in the absence of a "ground truth" it remains a major challenge to evaluate the impact of preprocessing choices on subsequent statistical analyses and results. Over time, preprocessing pipelines (i.e. a set of preprocessing steps) have become more complex and flexible, and  this increase in researcher degrees of freedom (termed multiverse analyses) has consistently been shown to affect the outcomes of neuroimaging studies \cite{carp2012plurality,norgaard2019optimization, botvinik2020variability}. 
The most common approach in the neuroimaging field is, to date, to use a single pipeline and ignore the heterogeneity of preprocessing choices. This approach not only makes abstraction of the multitude of possible results but is likely also sub-optimal because the best pipeline is more often than not, study, population or even subject dependent \cite{norgaard2019optimization, Churchill2015}. More concerning, neuroscientists might be tempted to “tune” the pipeline in order to obtain the most satisfying results. This will generally lead to spurious and non-reproducible results since the variability induced by the choice of pipeline is not independent from the results. However, since it is neither realistic nor optimal to move toward a single unified preprocessing pipeline, there is an urgent need for a statistical framework allowing to explore results among many preprocessing pipelines in a principled way.

The aim of this work is thus to provide a statistical framework that can aggregate the evidence from multiverse analyses to produce conclusions robust to the choice of the pipeline. More specifically, the present paper proposes a statistical sensitivity analysis providing:
\begin{enumerate}
    \item[(i)] visualizations of the heterogeneity of several preprocessing pipelines
    \item[(ii)] estimation of a global effect across all preprocessing pipelines
    \item[(iii)] quantification of the proportion of pipelines with evidence for an effect
    \item[(iv)] a statistical framework for testing hypotheses across pipelines such as “no effect across all pipelines”, “at least one pipeline with no effect”
\end{enumerate}

The corresponding software can be found at *anonymous*, including all code to reproduce all simulations and figures.

\section{Materials and experimental settings}

\subsection{Data}
\label{subsec:data}
We use two different data sources in our analyses: in sillico and real data.  For the in sillico data, different noise structures were chosen to reflect different configurations of pipelines and the sample size was varied to encompass small to larger scale clinical studies.
 In the real data analysis, we mimic how real neuroimaging studies compare an intervention to a reference measurement. Pipelines were selected independently of the intervention data using the healthy/placebo arm of \cite{frokjaer2015role} using results from \cite{norgaard2019optimization}. The proposed sensitivity analysis was illustrated on the intervention arm of the study where the follow-up value was compared to a reference value taken from a normative serotonergic atlas \cite{beliveau2017high}.

\bigskip

To simulate in sillico data, we consider the simple case of a single brain measurement (\(R=1\)), with a single binary exposure (\(P=1\)) following a Bernoulli distribution with parameter \(\pi=0.5\) (i.e. two balanced groups) and no covariates (\(C=0\)). Latent \(Y\) values for the brain measurement are simulated using a normal distribution with variance 1 and mean \(\beta\) times the exposure \(X\) values where \(\beta=0\) (null hypothesis) or \(\beta=0.5\) (alternative hypothesis). The observed \(Y\) is simulated for each pipeline adding pipeline specific noise to the latent \(Y\). This noise is simulated using a multivariate normal distribution with mean 0 and variance \(\Sigma_{\text{scenario 1}}\), \(\Sigma_{\text{scenario 2}}\), and \(\Sigma_{\text{scenario 3}}\) depending on the scenario. In scenario 1, we simulated many pipelines (\(J=20\)) with correlated homoscedastic noise, in scenario 2 a few pipelines (\(J=6\)) with uncorrelated heteroscedastic noise, and in scenario 3 many pipelines (\(J=20\)) with correlated heteroscedastic noise. Scenario 2 and 3 included one pipeline with high signal to noise ratio (SNR), i.e. low variance, and many pipelines with low SNR, i.e., high variance. The sample size was varied from \(n=10\) to \(n=500\) in each group, such that the smallest sample size was well below the number of parameters (\(2 \times J\) mean parameters, \(J^2\) variance-covariance parameters) in scenario 1 and 3 and the asymptotic regime was reached for the largest sample size. We generate 10,000 datasets per scenario and sample size - this provides sufficient precision about the mean, standard deviation, and rejection rate to neglect the Monte Carlo uncertainty. This data will be used  to assess the large sample size properties of the procedure (bias, relative efficacy, type 1 error control) in finite samples. 

\bigskip

To illustrate the use of the proposed sensitivity analysis on real data, we utilize neuroimaging results from a placebo-controlled, double-blinded, clinical study \cite{frokjaer2015role}. The study was registered and approved by the ethics committee for the capital region of Copenhagen (protocol-ID: H-2-2010-108) and registered as a clinical trial: \url{www.clinicaltrials.gov} under the trial ID NCT02661789. All subjects provided written informed consent prior to participation, in accordance with The Declaration of Helsinki II. The aim of the study was to assess the association between the emergence of depressive symptoms and change in cerebral serotonin transporter (SERT) availability following a hormonal treatment (\(p=1\)). Data is available from the CIMBI database (\url{www.cimbi.dk}) upon request. It consists of structural Magnetic Resonance Imaging (MRI) and Positron Emission Tomography (PET) imaging data for \(60\) healthy females who underwent a baseline scan, received either Placebo (\(n=30\)) or a GnRHa implant intervention (\(n=30\)), and participated in a follow-up scan. SERT availability estimates were extracted for each subject for 28 subcortical and cortical regions, and averaged across hemispheres, producing a final sample of \(R=14\) regions per subject and pipeline. These regions (amygdala, thalamus, putamen, caudate, anterior cingulate cortex, hippocampus, orbital frontal cortex, superior frontal cortex, occipital cortex, superior temporal gyrus, insula, inferior temporal gyrus, parietal cortex, and entorhinal cortex) were chosen because they cover the entire brain, and many are target regions in published serotonin transporter (SERT) PET studies. No covariates where consider  and (\(C=0\)).

\subsection{Data preprocessing}
Five preprocessing steps were used to curate the data and estimate the SERT availability (outcome measure). These steps include, motion correction (with/without), co-registration (4 options), delineation of volumes of interest (3 options), partial volume correction (4 options), and kinetic modeling for quantification of SERT availability (MRTM, SRTM, Non-invasive Logan and MRTM2). More information about the preprocessing choices can be found in \cite{norgaard2019optimization}. The combination of individual preprocessing steps leads to a number of \(J=2\times 3 \times 4^{3}=384\) possible combinations.  

\subsection{Notation and assumptions}
In the following section, we use generic notations to describe the proposed sensitivity analysis. We are interested in relating \(R\) brain measurements (\(\VY= \left(Y_1,\ldots,Y_R\right)\)) to \(P\) exposures or treatments  (\(\VX= \left(X_1,\ldots,X_P\right)\)) accounting for \(C\) covariates  (\(\VW= \left(W_1,\ldots,W_C\right)\)). We consider a set of \(J\) pipelines used to preprocess the neuroimaging data. For a given pipeline \(j \in \{1,\ldots,J\}\) we fit a statistical model with parameters \(\theta_j\) that relates \(\VY\) processed by pipeline \(j\), \(\VX\), and \(\VW\). We then obtain from this model an estimate \(\psiHat_k\) of the effect of interest (denoted \(\psi\)). In our real life example we use, for each pipeline, a paired t-test to compare the observed change in SERT availability to an atlas value so for \(j \in \{1,\ldots,J\}\), \(\widehat{\theta}_j=(\psiHat_j,\widehat{\sigma}^2_j)\) where \(\psiHat_j\) is the empirical mean and \(\sigma^2_j\) the empirical variance of the change in SERT availability (processed with pipeline \(j\)) between baseline of follow-up. 

\bigskip

We make the following working assumptions: first the observed data \(\left(\Sample_i\right)_{i\in\{1,\ldots\}}=\left(\Vy_i,\Vx_i,\Vw_i\right)_{i\in\{1,\ldots,n\}}\) correspond to independent and identically distributed replicates of \(\left(\VY,\VX,\VW\right)\). Second we have chosen a set of reasonable pipelines, meaning that the estimated effects \(\psiHat_1,\ldots,\psiHat_J\) found in the follow-up statistical analysis will converge to the right value \(\psi\) as the sample size increases. This set can include pipelines distorting the signal \(\VY\) (e.g. adding a fixed value) if that has no consequence, asymptotically, on the estimated effect (the mean change is not biased as the added value cancels out when substracting the baseline value to the follow-up value). Finally, when considering asymptotic results we will consider a fixed number of pipelines and let the sample size \(n\) increase to infinity.

\section{Proposed sensitivity analysis}
To be able to draw conclusions across pipelines, we not only need the result of each pipeline but also some information about how they relate. If all pipelines were equally reliable and equally different, we would weight each pipeline equally. If there exists one independent pipeline and a block of correlated pipelines, all equally reliable, then we would weight the independent pipeline more compared with other pipelines. By treating pipelines as black boxes, we can investigate their relation in terms of each observations influence on the estimated effects across pipelines, \(\VpsiHat=\left(\psiHat_1,\ldots,\psiHat_J\right)\). This relation is fully characterized by the joint distribution of the effects. Once estimated, we can extract summaries of this distribution, e.g. an average value, and carry out statistical tests, e.g. testing the compatibility between the observations and the joint distribution that would have been observed under a specific hypothesis.

\subsection{Estimating the joint distribution across pipelines}

The joint distribution could be obtained using a multivariate model, e.g. modelling data from all pipelines at once using a mixed model. Because of the complexity of the dependency among pipelines, this is, however, rarely feasible with the available sample size. Instead, and this matches common practice, we perform the same analysis separately for each pipeline and obtain a vector of estimated associations \(\VpsiHat\) with their standard errors \(\Vsigma_{\psiHat}=\left(\sigma_{\psiHat_1},\ldots,\sigma_{\psiHat_J}\right)\). Using tools from the semi-parametric theory (see \cite{kennedy2017semiparametric} and \cite{tsiatis2006semiparametric} for more details), we can approximate the influence of each observation on the estimate by a random variable called the influence function, denoted \(\varphi_{\psiHat_j}\) for pipeline \(j\), and satisfying:
 \begin{align*}
 \sqrt{n}\left(\psiHat_j-\psi\right)=\frac{1}{\sqrt{n}} \sum_{i=1}^n \varphi_{\psiHat_j}(\Sample_i) + o_p(1)
 \end{align*}
 where \(o_p(1)\) denotes a residual term that convergences toward zero in probability as the sample size tends to infinity. As shown in appendix \ref{SM:IF}, \(\varphi_{\psiHat_j}\) has a simple expression when \(\psiHat_j\) is the empirical mean or an element of a maximum likelihood (ML) estimator. Since this decomposition applies to all pipelines, we get from the multivariate central limit theorem that the joint distribution of the estimates is asymptotically multivariate normal. It has mean \(\psi\) and its variance-covariance, denoted \(\Sigma_{\VpsiHat}\), is the same as the one of \(\Vvarphi_{\VpsiHat}=\left(\varphi_{\psiHat_1},\ldots,\varphi_{\psiHat_J}\right)\) divided by \(n\). Note that with limited number of observations, typically when \(n<J\), the estimated variance-covariance \(\SigmaHat_{\VpsiHat}\) based on the estimated influence function is not guaranteed to be positive definite.

\subsection{Testing hypotheses across pipelines}
The global null hypothesis “no effect across all pipelines” can be tested using a max-test approach: more extreme realizations would correspond to larger values of the maximum statistic \(\VtHat_{\max}=\max(|\tHat_1|,..,|\tHat_J|)\) where \(|.|\) denotes the absolute value and \(\tHat_j=\frac{\psiHat_j}{\sigma_{\psiHat_j}}\). A p-value may therefore be computed by integrating the joint density under the null hypothesis outside of the domain \(\mathcal{D}(\VtHat_{\max})=[-\VtHat_{max},\VtHat_{max}]^{\otimes J}\) (see figure~\ref{fig:maxT} in appendix \ref{SM:bivGaus}). Here, we use the notation \([a,b]^{\otimes J}=\prod_{j=1}^J [a,b]\) that represents the Cartesian product between \(J\) intervals \([a,b]\). The value \(t_c\), such that the integral outside \(\mathcal{D}(t_c)\) equals \(\alpha\), provides a critical threshold for the estimated test statistics \(\left(|\tHat_1|,..,|\tHat_J|\right)\). This threshold can also be used to derive confidence intervals. 

\bigskip

The null hypothesis “at least one pipeline with no effect” is an intersection union test. As such it can be rejected if and only if all the un-adjusted p-values relative to each pipeline are below \(\alpha\) or equivalently if the largest un-adjusted p-value is below \(\alpha\). 

\bigskip

The proportion of pipelines where there is evidence for an effect \(\eta\) can be estimated as \(\tilde{\eta} = \frac{1}{J}\sum_{j=1}^{J } \Ind_{\left|\tHat_j\right|<t_c}\) where \(\Ind_{.}\) denotes the indicator function. One drawback with this non-parametric estimator is that it is a non-smooth function of \(\tHat_j\), making the associated uncertainty difficult to evaluate. Instead, one can use a parametric approach, assuming normally distributed test statistics:
\begin{align*}
 \etaHat=\frac{1}{J}\sum_{j=1}^{J} 1-\Prob\left[-t_c< \tHat_j < t_c\right] = 1-\frac{1}{J}\sum_{j=1}^{J} \Phi\left(t_c - \tHat_j \right) - \Phi\left(-t_c - \tHat_j \right)
\end{align*}
which is a smooth (but complex) function of the model parameters. Here \(\Phi\) refers to the cumulative distribution function of a standard normal distribution. The uncertainty about the estimator can therefore be derived using a non-parametric bootstrap or a delta method where \(Var(\etaHat) = \dpartial{\etaHat}{\Theta} \Sigma_{\ThetaHat} \trans{\dpartial{\etaHat}{\Theta}} \) where \(\Theta =\left(\theta_j\right)_{j \in \{1,\ldots,J\}}\) is the set of parameters of the statistical model across pipelines.

\subsection{Visualizing the heterogeneity across pipelines}
We have derived that the estimated associations are, asymptotically, normally distributed. They can therefore be summarized by their expectation and variance-covariance matrix (i.e. standard errors and correlation matrix). We suggest two graphical displays to visualize the heterogeneity of the results across pipelines:
\begin{itemize}
    \item A heatmap of the estimated correlation among estimates, obtained by converting \(\SigmaHat_{\VpsiHat}\) into a correlation matrix. \footnote{Re-ordering the pipelines may be useful to better visualize blocks of pipelines that are especially correlated. This can for instance be performed by converting R to a dissimilarity matrix and then use hierarchical clustering.}
    \item A forest plot displaying \(\left(\psiHat_1,\ldots,\psiHat_J\right)\) and \(\left(\sigmaHat_{\psiHat_1},\ldots,\sigmaHat_{\psiHat_J}\right)\) through the confidence intervals, possibly using the previously established re-ordering.
\end{itemize}

\subsection{Estimating a global effect across pipelines}
Several methods can be used for estimating a global effect across pipelines. A naive method would be to compute the mean of the estimated associations:
\begin{align*}
 \PsiHat_{\text{average}}=\frac{1}{J}\sum_{j=1}^{J} \psiHat_j
\end{align*}
This estimator will, however, not be efficient if some pipelines lead to more precise estimates, i.e. \(\left(\sigma_{\psiHat_1},\ldots,\sigma_{\psiHat_J}\right)\) are not equal. Intuitively, we would like to pool the estimates with weights inversely proportional to the standard errors such that we put more weight on precise estimates:
\begin{align*}
 \PsiHat_{\text{pool-se}}=\sum_{j=1}^{J} w^{se}_j \psiHat_j \text{ where }  w^{se}_j=\frac{1/\sigma^2_{\psiHat_j}}{\sum_{j=1}^J 1/\sigma^2_{\psiHat_j}}
\end{align*}
However, we also need to take into account the correlation between estimates. Indeed, perfectly correlated estimates should weight as if there was only one estimate. To do so, we can use the following GLS estimator of the global effect:
\begin{align}
 \PsiHat_{\text{GLS}}= \left(\trans{\Vone} \SigmaHat_{\PsiHat}^{-1}\Vone\right)^{-1} \trans{\Vone} \SigmaHat_{\PsiHat}^{-1} \VpsiHat =\sum_{j=1}^{J} w^{GLS}_j \psiHat_j \label{eq:GLS-Inv}
\end{align}
where \(\Vone\) is a column vector filled with ones. Equation \ref{eq:GLS-Inv} can be shown to be equivalent to performing a spectral decomposition of \(\SigmaHat_{\VpsiHat}\), and use the eigenvectors to combine estimates into independent components that can be pooled according to weights proportional to the eigenvalues (appendix \ref{SM:GLSPCA}). This is used when \(\Sigma_{\VpsiHat}\) is singular to compute \(\PsiHat_{\text{GLS}}\), by  restricting the spectral decomposition to the eigenvalues above a given threshold (\(\epsilon = 10^{-10}\)). In the simple case where \(R=1\), \(p=1\), \(C=0\), brain measurements are jointly normally distributed, \(X\) is binary, and there is no missing data, the GLS estimator can be shown to be asymptotically efficient (appendix \ref{SM:GLSefficient}). 

\bigskip

Cox and colleagues \cite{cox2006generalized} studied a similar estimator when \(J=2\) and found that, under unequal variance, the global estimate can be outside of the range of the (pipeline specific) estimates. As a remedy to this unpleasant behavior, we propose a constrained GLS estimator, denoted \(\PsiHat_{\text{constrained GLS}}\), which constrains the weight of each estimate to be at most 1 in absolute value and ensures that the sum of the weights is 1:
\begin{align*}
w_j^{\text{constrained GLS}} = \frac{w_j^{GLS}}{\kappa + \max_{j \in  \{1,\ldots, J\}} |w_j^{GLS}|} + \frac{1}{J}\left(1-\frac{1}{\kappa + \max_{j \in  \{1,\ldots, J\}} |w_j^{GLS}|} \right) 
\end{align*}

where \(\kappa\) is chosen to satisfy the constraints. Note that if some of the pipelines may induce some bias, the previous approaches will propagate this bias and therefore be unsatisfactory. Systematic differences between pipelines can be investigated by comparing the estimates between pipelines, e.g. \(\psiHat_j-\psiHat_{j^\prime}\) and using \(\SigmaHat_{\VpsiHat}\) to obtain the corresponding uncertainty \(\Var\left[\psiHat_j-\psiHat_{j^\prime}\right]=\Var\left[\psiHat_j\right]+Var\left[\psiHat_{j^\prime}\right]-2 \Cov\left(\psiHat_j,\psiHat_{j^\prime}\right)\). 

\section{Results}
\subsection{Simulation results - pooled estimator}
We compare the performance of the proposed estimators when using unbiased pipelines on simulated data. For each dataset \(\Sigma_{\text{scenario 1}}\), \(\Sigma_{\text{scenario 2}}\), and \(\Sigma_{\text{scenario 3}}\), we computed the four previously described estimators of the global effect, \(\PsiHat_{\text{average}}\), \(\PsiHat_{\text{pool-se}}\), \(\PsiHat_{\text{GLS}}\), and \(\PsiHat_{\text{constrained GLS}}\). P-values for \(\PsiHat_{\text{pool-se}}\), \(\PsiHat_{\text{GLS}}\), and \(\PsiHat_{\text{constrained GLS}}\) were computed neglecting the uncertainty of the weights (\(w_j^{\text{se}}\),  \(w_j^{\text{GLS}}\),  \(w_j^{\text{constrained GLS}}\)). 

\bigskip

\begin{figure}[!t]
\centering
\includegraphics[width = \textwidth]{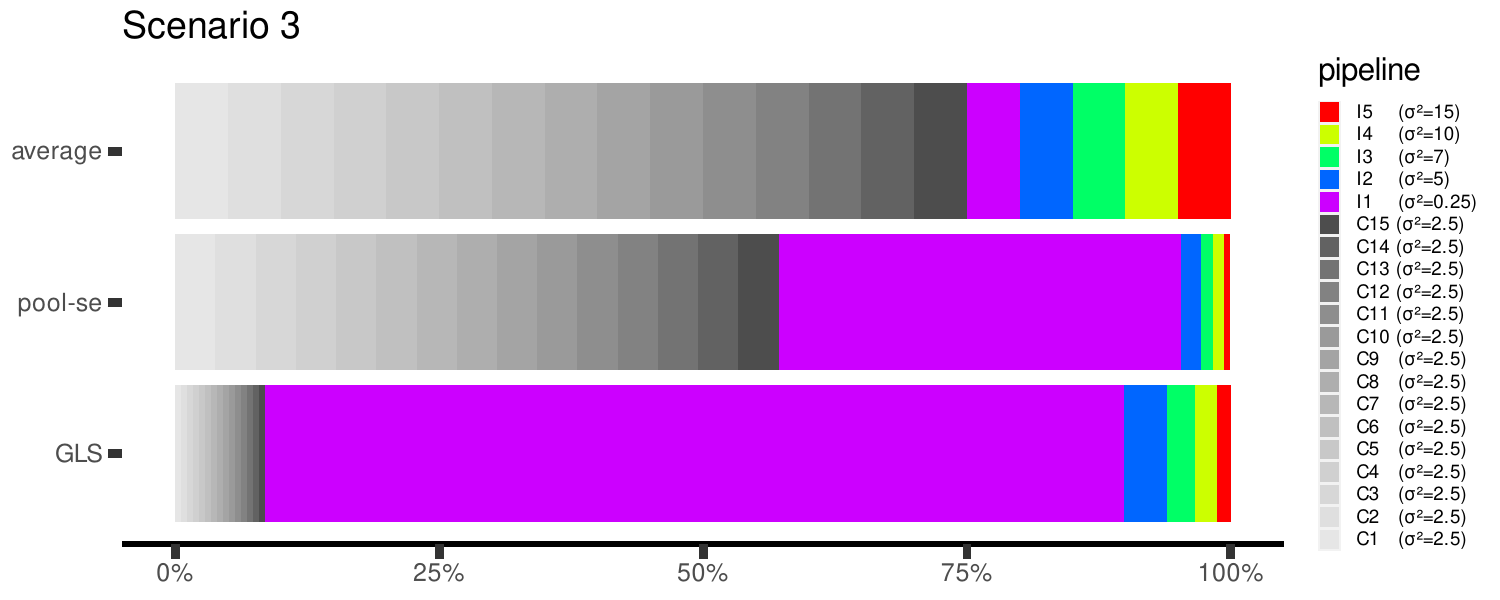}
\caption{Large sample weights used by \(\PsiHat_{\text{average}}\), \(\PsiHat_{\text{pool-se}}\), and the GLS estimators (\(\PsiHat_{\text{GLS}}\) or \(\PsiHat_{\text{constrained GLS}}\)) to combine the pipeline specific estimates in scenario 3. Weights relative to the correlated pipeline are shown in shades of gray (first fifteen blocks); weights relative to the independent pipeline are shown using rainbow colors (last five blocks). In the legend the variance for a given pipeline effect is indicated in parenthesis.}
\label{fig:sim-estimator-weights}
\end{figure}

\noindent \textbf{Weights}: Based on \(\Sigma_{\text{scenario 1}}\), \(\Sigma_{\text{scenario 2}}\), and \(\Sigma_{\text{scenario 3}}\), we can compute how \(\PsiHat_{\text{average}}\), \(\PsiHat_{\text{pool-se}}\), and \(\PsiHat_{\text{GLS}}\) (or \(\PsiHat_{\text{constrained GLS}}\)) would weight the results from each pipeline if the variance-covariance matrix of the pipeline estimates was known (Figure \ref{fig:sim-estimator-weights}). In scenario 1, \(\PsiHat_{\text{average}}\) and \(\PsiHat_{\text{pool-se}}\) would provide equal weight to all pipelines with a weight of 5\%, while \(\PsiHat_{\text{GLS}}\)
would weight by 2\% the correlated pipelines and by 14\% the uncorrelated pipelines (in this paragraph all weights are rounded for readability). In scenario 2, \(\PsiHat_{\text{average}}\) equally weights all pipelines by 17\% while \(\PsiHat_{\text{pool-se}}\) and \(\PsiHat_{\text{GLS}}\) would use the following weights: 8\%, 82\%,  4\%,  3\%,  2\%, 1\%, favoring the high SNR pipeline. In scenario 3, \(\PsiHat_{\text{average}}\) would equally weight all pipelines by 5\%, \(\PsiHat_{\text{pool-se}}\) would weight each correlated pipeline by 3.8\% and the remaining pipelines by 38\% (high SNR pipeline),  2\%,  1.3\%,  1\%,  0.6\%, while \(\PsiHat_{\text{GLS}}\) would weight by 0.6\% the correlated pipelines and by 81\% (high SNR pipeline),  4\%,  3\%,  2\%,  1\% the remaining pipelines.

\bigskip

\noindent \textbf{Estimate and bias}: No significant bias was found. Under the null, the proportion of pipelines with evidence for an effect ranged between 5\% and 7\% and was relatively stable with sample size. Under the alternative, the proportion of pipelines increased from 6.3\% in scenario 1 to 98\% in scenario 1 when the sample size increased from \(n=20\) to \(n=500\). A similar behavior was observed in the other scenarios (up to 63\% in scenario 2 and up to 79\% in scenario 3). \newline
\textcolor{red}{TODO: add random intercept model}

\medskip

\begin{figure}[!h]
\centering
\includegraphics[width = \textwidth]{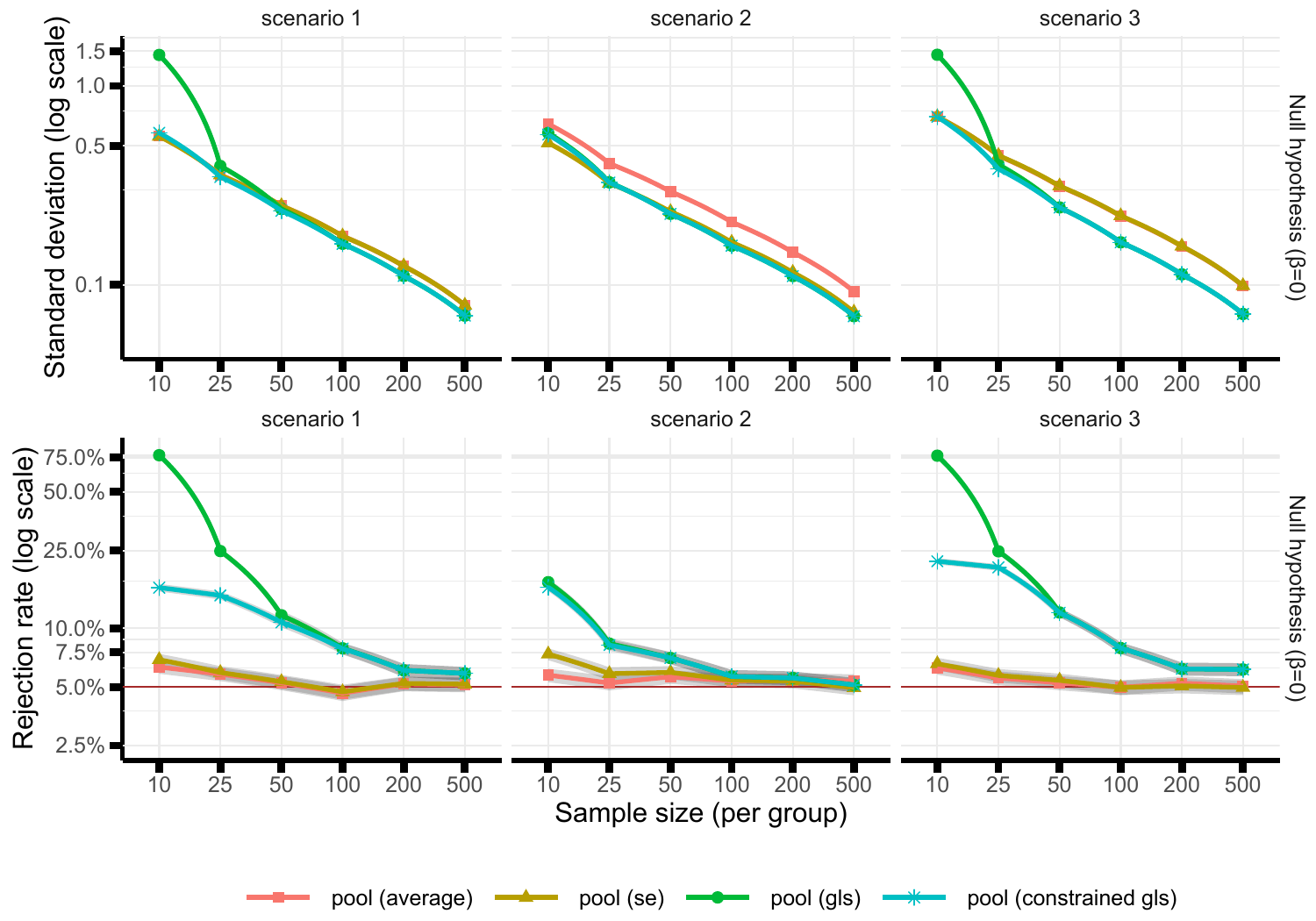}
\caption{\textbf{Upper panel:} empirical standard deviation of the estimated common effect for each estimator, scenario, and sample size under the null hypothesis. For large samples the green line (GLS) is covered by the blue line
(constrained GLS) in all scenarios. In scenario 1 and 3, the red line (pool-average) is covered by the yellow line (pool se). In scenario 1, the GLS estimators have the same or higher standard deviation as the average for small samples and lower standard deviation (approx. 12\% lower) for large samples. \newline \hphantom{Fig. X} \textbf{Lower panel:} rejection rate under the null hypothesis (i.e. type 1 error). Shaded area represents the Monte Carlo uncertainty.}
\label{fig:ressim-eff}
\end{figure}

\noindent \textbf{Efficiency}: Similar results were obtained for \(\beta=0\) and \(\beta=0.5\), so we will only discuss the former case (Figure \ref{fig:ressim-eff}). In scenario 1, \(\PsiHat_{\text{average}}\) and \(\PsiHat_{\text{pool-se}}\) showed similar empirical variance, whereas \(\PsiHat_{\text{GLS}}\) showed higher empirical variance in very small samples (+159\% for \(n=10\), +13\% for \(n=25\)) and lower empirical variance at larger sample sizes (e.g. -12\% for \(n=500\)). In scenario 2, \(\PsiHat_{\text{pool-se}}\) and \(\PsiHat_{\text{GLS}}\) showed similar empirical variance (slightly higher for \(\PsiHat_{\text{GLS}}\) in low sample sizes and slightly lower afterwards), both smaller compared to  \(\PsiHat_{\text{average}}\), e.g. 10\% for \(n=10\) and -24\% for \(n=500\) for the GLS estimator. Results in scenario 3 were similar to scenario 1 up to a larger decrease in variance in large samples for GLS estimators (-28\% for n=500). \(\PsiHat_{\text{constrained GLS}}\) was similar to \(\PsiHat_{\text{GLS}}\)  but with better small sample properties (at most +4\% in standard error compared to \(\PsiHat_{\text{average}}\)). 

\medskip

\noindent \textbf{Type 1 error}: Across all scenarios, the type 1 errors for \(\PsiHat_{\text{average}}\) and \(\PsiHat_{\text{pool-se}}\) were well controlled except in very small samples where small deviations from the nominal level were observed (maximum of 7\% for \(\PsiHat_{\text{average}}\) and 8\% for \(\PsiHat_{\text{pool-se}}\)). When neglecting the uncertainty about the weights, the type 1 error control of \(\PsiHat_{\text{GLS}}\) and \(\PsiHat_{\text{constrained GLS}}\) were only controlled for large samples, i.e. for \(n = 500\); large type 1 error rates were found in very small samples (>10\%). \\

\subsection{Real-world application}

\begin{figure}[!b]
\centering
\includegraphics[width = \textwidth]{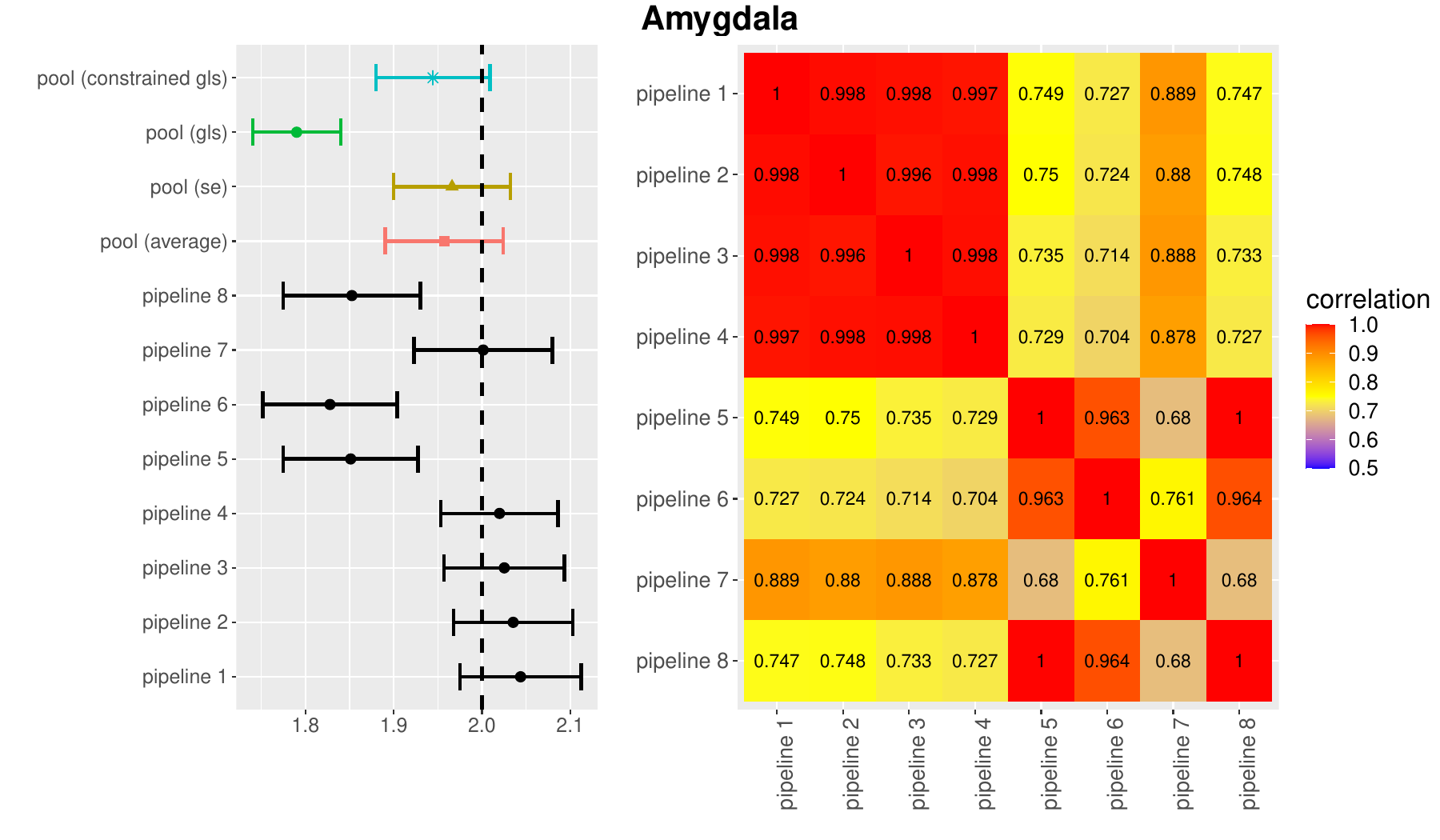}
\caption{\textbf{Left panel:} forest plot of the estimated SERT availability in the amygdala for the intervention group (point and full line) vs. the normative values (dashed line) for each pipeline and the four proposed pooled estimators. \newline \hphantom{Fig. X} \textbf{Right panel:} correlation of the estimated SERT availability between pipelines. \newline Pipeline 1: with motion correction (MC), boundary based registration (BBR) using the time-weighted average PET image (twa), and MRTM2 as kinetic modeling choice. \newline Pipeline 2: MC, normalized mutual information registration (NMI) using twa, MRTM2.  \newline Pipeline 3: MC, BBR using the average PET image, and MRTM2.  \newline  Pipeline 4: MC, NMI using the average PET image, and MRTM2.  \newline Pipeline 5: MC, BBR\_twa, and MRTM.  \newline Pipeline 6: no motion correction (nMC), BBR\_twa, and MRTM.  \newline Pipeline 7: nMC, BBR\_twa, and MRTM2. \newline Pipeline 8: MC, BBR\_twa, and SRTM.}
\label{fig:ressim-bias}
\end{figure}

We illustrate the statistical sensitivity analysis with data described in section \ref{subsec:data} in which the SERT availability was assessed after a drug intervention and compared to normative values. 
We start by studying the behavior of the four statistical estimators (pooled GLS, pooled constrained GLS, pooled average and pooled SE) to estimate a common effect across pipelines for the given null hypothesis, reported as a forest plot in Figure \ref{fig:ressim-bias} (left panel). 
The dashed vertical line in Figure \ref{fig:ressim-bias} represents the normative value, and all horizontal error bars represent the estimated effect (mean and 95\% CI) for a given pipeline and estimator. 
Across a reasonable set of preprocessing pipelines, 3 of the 8 selected reject the null hypothesis (as indicated by the non-overlapping CI with the normative value) with estimated percent differences between groups ranging between -9\% (pipeline 6) and 2.5\% (pipeline 1). 
The pooled constrained GLS, pooled SE and pooled average, all fail to reject a common effect across pipelines, whereas the GLS estimator rejects the null hypothesis. 
However, when inspecting the pipeline weights for the GLS estimator for this data, it assigned a very high weight to four pipelines (i.e. weight above 1 in absolute values), leading to an unreliable estimate as illustrated by the large standard deviation found in the simulation study for low sample sizes (Figure \ref{fig:ressim-eff}). The constrained GLS estimator did not exhibit this problem and had weights between -0.79 and 1.  

\bigskip

\indent Figure \ref{fig:ressim-bias} (right panel) shows a heatmap for the estimated correlations across preprocessing pipelines, ranging from 0.68 to 1. Pipelines 1-4 show a very high correlation with each other (only varying the registration choices), whereas pipelines 5, 6 and 8 are equally correlated with each other (different kinetic models), and pipeline 7 is somewhat in between (no motion correction). The heatmap captures important differences in the correlation structure between pipelines, suggesting that not all pipelines perform similarly with moderate levels of unexplained variance. Pipelines 4 to 8 exhibit numerically smaller correlation compared to pipelines 1 to 4 (and similar variance), which explains why GLS estimators produce estimates with lower numerical values compared to pooled estimators ignoring the correlation, as GLS estimators assign more weight to the last four pipelines.

\bigskip

\indent Notably, no correction for multiple comparisons was carried out across regions and pipelines at this point. The rationale for not including this (as should otherwise always be carried out) is that we wanted to make our analysis as comparable as possible to the Neuroimaging Analysis Replication and Prediction Study \cite{botvinik2020variability}, where each participating institution analyzed the data using their own established pipeline and tested only a single region in a hypothesis-driven fashion.

\section{Discussion}
Looking first at the simulated data results, we observe that, asymptotically, \(\PsiHat_{\text{GLS}}\) is the best estimator, leading to more precise estimates than  \(\PsiHat_{\text{average}}\) or  \(\PsiHat_{\text{pool-se}}\). It is, however, not suited for small sample sizes or large numbers of pipelines, as it exhibits large variance. We proposed an alternative estimator \(\PsiHat_{\text{constrained GLS}}\) which can be seen as regularizing the GLS estimator toward the empirical average in small samples. Even though this considerably improves the small sample performance of the estimator, other regularization approaches may lead to further gain. An alternative approach could be to regularize the estimated variance-covariance matrix between pipeline specific estimates, e.g. using graphical lasso. However, this is challenging since the correlation structure among pipelines is typically complex and non-sparse. The proposed method to quantify the uncertainty of \(\PsiHat_{\text{constrained GLS}}\) is numerically fast, but unreliable in small samples or with a large number of pipelines. However, these latter points are beyond the scope of this paper, and is left for future work. 
In the real-world application, we observed that all estimators could be readily applied, and three of them performed as expected based on the results from individual pipelines. Only the \(\PsiHat_{\text{GLS}}\) estimator performed differently and was the only one rejecting the null hypothesis hinting at an effect across pipelines. This is though, due to the estimator not being able to fit the weights properly in small sample sizes. Since PET neuroimaging studies are rarely beyond sample sizes of n>50, other estimators should be used. We recommend using \(\PsiHat_{\text{pool-se}}\) or \(\PsiHat_{\text{constrained GLS}}\): the latter  when pipelines are not similarly related and the sample size is moderate to large, otherwise the former.

\bigskip

The framework that we are proposing is not without limitations. The underlying assumption of combining results across pipelines in our analysis is that all pipelines are unbiased. This can however not always be guaranteed. Alternative approaches (e.g. STAPLE \citep{Warfield2004}) could be used to reduce the bias of the pooled estimators by assuming a majority of unbiased pipelines or identifying clusters of pipelines and pooling pipeline-specific estimates within clusters.

\section{Conclusion}
In this work, we have developed a statistical sensitivity analysis that can quantify the impact of different preprocessing choices on subsequent statistical analyses. As has been reported in previous studies, we observe that the influence of preprocessing pipelines on subsequent statistical analysis can be quite large. Hence, we provide tools for statistical analyses that can aggregate multiple analyses of the same data. We introduce four statistical estimators, \(\PsiHat_{\text{average}}\), \(\PsiHat_{\text{pool-se}}\), \(\PsiHat_{\text{GLS}}\), and \(\PsiHat_{\text{constrained GLS}}\) to combine the pipeline specific estimates. This enables testing hypotheses across pipelines, such as “no effect across all pipelines” or “at least one pipeline with no effect”. The proposed framework is generic and can be applied to any imaging modality.
%
%
\bibliographystyle{splncs04}
\bibliography{bibliography.bib}

\begin{thebibliography}{10}
\providecommand{\url}[1]{\texttt{#1}}
\providecommand{\urlprefix}{URL }
\providecommand{\doi}[1]{https://doi.org/#1}

\bibitem{beliveau2017high}
Beliveau, V., et~al.: A high-resolution in vivo atlas of the human brain's
  serotonin system. Journal of Neuroscience  \textbf{37}(1),  120--128 (2017)

\bibitem{botvinik2020variability}
Botvinik-Nezer, R., et~al.: Variability in the analysis of a single
  neuroimaging dataset by many teams. Nature  \textbf{582}(7810),  84--88
  (2020)

\bibitem{carp2012plurality}
Carp, J.: On the plurality of (methodological) worlds: estimating the analytic
  flexibility of f{MRI} experiments. Frontiers in neuroscience  \textbf{6},
  ~149 (2012)

\bibitem{Churchill2015}
Churchill, N.W., et~al.: {An automated, adaptive framework for optimizing
  preprocessing pipelines in task-based functional MRI}. PLoS ONE
  \textbf{10}(7),  1--25 (2015)

\bibitem{cox2006generalized}
Cox, M., Ei{\o}, C., Mana, G., Pennecchi, F.: The generalized weighted mean of
  correlated quantities. Metrologia  \textbf{43}(4), ~S268 (2006)

\bibitem{frokjaer2015role}
Frokjaer, V.G., et~al.: Role of serotonin transporter changes in depressive
  responses to sex-steroid hormone manipulation: a positron emission tomography
  study. Biological psychiatry  \textbf{78}(8),  534--543 (2015)

\bibitem{kennedy2017semiparametric}
Kennedy, E.H.: Semiparametric theory. arXiv:1709.06418  (2017)

\bibitem{norgaard2019optimization}
N{\o}rgaard, M., , et~al.: Optimization of preprocessing strategies in positron
  emission tomography ({PET}) neuroimaging: a {[11C]} {DASB PET} study.
  Neuroimage  \textbf{199},  466--479 (2019)

\bibitem{norgaard2019preprocessing}
N{\o}rgaard, M., Ozenne, B., Svarer, C., Frokjaer, V.G., Schain, M., Strother,
  S.C., Ganz, M.: Preprocessing, prediction and significance: Framework and
  application to brain imaging. International Conference on Medical Image
  Computing and Computer-Assisted Intervention pp. 196--204 (2019)

\bibitem{Poldrack2020}
Poldrack, R.A., et~al.: {Establishment of Best Practices for Evidence for
  Prediction: A Review}. JAMA Psychiatry  \textbf{77}(5),  534--540 (2020)

\bibitem{stenbaek2017brain}
Stenb{\ae}k, D.S., Fisher, P.M., Ozenne, B., Andersen, E., Hjordt, L.V.,
  McMahon, B., Hasselbalch, S.G., Frokjaer, V.G., Knudsen, G.M.: Brain
  serotonin 4 receptor binding is inversely associated with verbal memory
  recall. Brain and behavior  \textbf{7}(4),  e00674 (2017)

\bibitem{tsiatis2006semiparametric}
Tsiatis, A.A.: Semiparametric theory and missing data. Springer (2006)

\bibitem{van2000asymptotic}
Van~der Vaart, A.W.: Asymptotic statistics, vol.~3. Cambridge university press
  (2000)

\bibitem{Warfield2004}
Warfield, S.K., Zou, K.H., Wells, W.M.: {Simultaneous truth and performance
  level estimation (STAPLE): An algorithm for the validation of image
  segmentation}. IEEE Transactions on Medical Imaging  \textbf{23}(7),
  903--921 (2004). \doi{10.1109/TMI.2004.828354}

\end{thebibliography}

\clearpage
 \appendix

 \renewcommand{\thefigure}{\Alph{figure}}
 \renewcommand{\thetable}{\Alph{table}}
 \renewcommand{\theequation}{\Alph{equation}}

 \setcounter{figure}{0}    
 \setcounter{table}{0}    
 \setcounter{equation}{0}

 \clearpage

\begin{subappendices}
\renewcommand{\thesection}{\Alph{section}}

\section{Influence function}
\label{SM:IF}

In the univariate case (i.e. \(R=P=1\)) without covariate (\(C=0\)), the effect of interest \(\psi\) may be the Pearson’s correlation coefficient:
\begin{align*}
\psi=\frac{\Esp[XY]-\Esp[X]\Esp[Y]}{\sqrt{\Esp[X^2]-E[X]^2}\sqrt{\Esp[Y^2]-\Esp[Y]^2} }
\end{align*}
when X is continuous or the mean difference when X is binary.
\begin{align*}
\psi=\Esp[Y|X=1]-\Esp[Y|X=0] 
\end{align*}
Here \(\Esp[.]\) denotes the expectation and \(\Esp[.|.]\) the conditional expectation. The later case leads to simple expression: \(\psiHat_j\) is the empirical mean difference between the groups
\begin{align*}
\psiHat_j=\frac{1}{n} \sum_{i=1}^n \left(\frac{y_{ij}x_{i}}{\pi}-\frac{y_{ij}(1-x_{i})}{1-\pi}\right)    
\end{align*}
where \(\pi\) denotes the proportion of observations with \(X=1\), \(y_{ij}\) the brain signal for individual \(i\) processed by pipeline \(j\), and \(x_{i}\) the exposure value for individual \(i\). The previous expression is equivalent to:
\begin{align*}
\sqrt{n}\left(\psiHat_j-\psi_j\right)=\frac{1}{\sqrt{n}} \sum_{i=1}^n \varphi_{\psiHat_j}(\Sample_i)  
\end{align*}
where \(\varphi_{\psiHat_j}(\Sample_i)=\frac{y_{ij}x_i}{\pi}-\frac{y_{ij}(1-x_i)}{1-\pi} -\psi_j\), \(\Sample_j=\left(y_{i1},\ldots,y_{iJ},x_i\right)\), and \(\psi_j\) is the large sample value of \(\psiHat_j\). Denoting \(Y_j\) the random variable representing the brain signal processed by pipeline \(j\), the estimator can be seen as the empirical average of independent realizations of a new random variable \(\frac{Y_{j}X}{\pi}-\frac{Y_{j}(1-X)}{1-\pi} -\psi_j\), called the influence function of \(\psiHat_j\). Thus from the multivariate central limit theorem we get that the joint distribution of the estimates is asymptotically multivariate normal. It has mean \(\psi\) (under our assumption that \(\forall j\in\{1,\ldots,J\}, \psi=\psi_j\)) and its variance-covariance, denoted \(\Sigma_{\VpsiHat}\), is the same as the one of \(\Vvarphi_{\VpsiHat}=\left(\varphi_{\psiHat_1},\ldots,\varphi_{\psiHat_J}\right)\) divided by \(n\). Because \(\Vvarphi_{\VpsiHat}\) involves some unknown parameters like \(\pi\) and \(\psi_j\) we do not observe it and cannot directly estimate \(\Sigma_{\VpsiHat}\). However, by plugging our estimates of these unknown parameters we can approximate \(\varphi_{\psiHat_j}\) as \(\varphiHat_{\psiHat_j}(\Sample_{ij})=\frac{y_{ij}x_{i}}{\frac{1}{n} \sum_{i=1}^n x_{i}}-\frac{y_{ij}(1-x_{i})}{1-\frac{1}{n}\sum_{i=1}^n x_{i}} - \frac{1}{n} \sum_{i=1}^n \left(\frac{y_{ij}x_{i}}{\pi}-\frac{y_{ij}(1-x_{i})}{1-\pi}\right)\) and approximate \(\Sigma_{\VpsiHat}\).

\bigskip

In a more general case, we would define a statistical model \(\Model(\Theta)\) relating \(\VX\) and \(\VY\) via a parameter \(\psi\). \(\psi\)  may be an element of \(\Theta\), the set of model parameters, or a function of elements of \(\Theta\). For instance one could use a latent variable model (LVM) with two latent variables, one summarizing the brain measurements and another summarizing the exposure variables. \(\psi\) is then the coefficient relating the two latent variables. See Figure 1 of \cite{stenbaek2017brain} for a graphical representation of a LVM - in this example the latent variable “LVu” represents the PET measurement and the latent variable “LVpos” the memory relative to positive word. In this more general case the previous decomposition does not hold exactly but up to a residual term \(o_p(1)\) which converges to 0 as the sample approaches infinity:
\begin{align*}
\sqrt{n}\left(\psiHat_j-\psi_j\right)=\frac{1}{\sqrt{n}} \sum_{i=1}^n \varphi_{\psiHat_j}(\Sample_i) + o_p(1) \end{align*}
as indicated in the main text of this article. This decomposition exists for any estimator \(\psiHat\) derived from an M-estimator \cite{van2000asymptotic} (section 5.3), including likelihood-based estimators. Denote by \(\widehat{\theta}_j\) the ML estimator and \(\PsiHat_j = \trans{c} \widehat{\theta}_j \) the parameter of interest (\(c\) may be a vector starting by 1 and followed by 0's, i.e. selects the first element of \(\widehat{\theta}_j\). The corresponding influence function only involves the first two derivatives of the log-likelihood (formula 3.6 in \cite{tsiatis2006semiparametric}):
\begin{align*}
\varphi_{\psiHat_j}(\Sample_i) = - \trans{c}\Esp\left[ \frac{\partial \Score_j(\Sample_i,\theta_j)}{\partial \theta_j} \right]^{-1} \Score_j(\Sample_i,\theta_j) 
\end{align*}
with \(\Score_j(\Sample_i,\theta_j)\) being the score for individual \(i\) when considering pipeline \(j\), i.e. vector containing the first derivatives of the log-likelihood contribution of individual \(i\). Once the influence function has been estimated for each individual, we can use it to obtain a consistent estimator of \(\Sigma_{\VpsiHat}\):
\begin{align*}
\SigmaHat_{\VpsiHat}=\frac{1}{n} \sum_{i=1}^n \trans{\Vvarphi_{\VpsiHat}(\Sample_{i})} \Vvarphi_{\VpsiHat}(\Sample_{i})
\end{align*}
where \(\trans{\Vvarphi_{\VpsiHat}(\Sample_{i})}\) denotes the transpose of the \(J\)-dimensional vector of influence functions relative to individual \(i\). Practically speaking the above allows us assess the variance-covariance matrix of effects across pipelines, e.g., if pipelines are completely independent in terms of their estimated effect, then \(\SigmaHat_{\VpsiHat}\) would be a diagonal matrix with the uncertainty of the estimated effect per pipeline in the diagonal. However, if the estimated effects across pipelines are correlated, the matrix would not be sparse.

\clearpage

\section{Integration of the Gaussian density - bivariate case}
\label{SM:bivGaus}

\begin{figure}[!tbhp]
\centering
\includegraphics[width = 8cm]{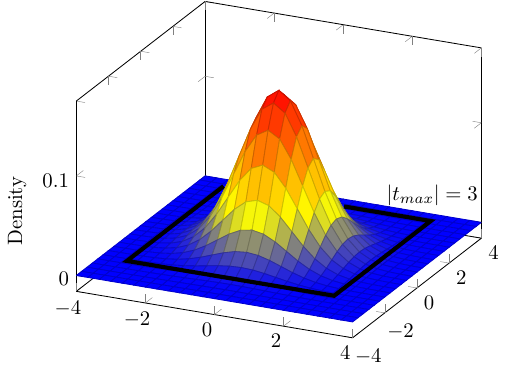}
\caption{Density of the two dimensional standard normal distribution (colored surface). The black line delimits the domain \(\mathcal{D}(\VtHat_{\max})\) where \(J=2\), \(\tHat_1=1.5\), \(\tHat_2=3\), and \(\Sigma_{\VpsiHat}\) is the identity matrix with two rows and two columns. The area under the blue surface external to the black line corresponds to the p-value relative to the first test, adjusted for two tests.}
\label{fig:maxT}
\end{figure}

\clearpage

\section{Reformulation of the GLS estimator}

\subsection{Via a spectral decomposition}
\label{SM:GLSPCA}

\(\SigmaHat_{\psiHat}\) being a symmetric semi-positive definite matrix, it admits the decomposition \(\SigmaHat_{\psiHat}=QD\trans{Q}\) where \(Q\) is an orthogonal matrix (i.e. \(Q\trans{Q}=I_J\), the identity matrix) and \(D\) is a diagonal matrix with non negative values \(\left(\lambda_1,\ldots,\lambda_J\right)\). Thus:
\begin{align*}
\VpsiHat_{\text{GLS}}
= \left(\trans{\Vone} \SigmaHat_{\PsiHat}^{-1}\Vone\right)^{-1} \trans{\Vone} \SigmaHat_{\PsiHat}^{-1} \VpsiHat
= \left(\trans{\Vone} Q D^{-1} \trans{Q}\Vone\right)^{-1} \trans{\Vone} Q D^{-1} \trans{Q} \VpsiHat
\end{align*}
We first note that \(\bar{Q}=\trans{\Vone}Q\) is a vector with element \(q\) equal to the sum (column-wise) of the eigenvectors.  Then:
\begin{align*}
\trans{\Vone} Q D^{-1} \trans{Q}\Vone=\sum_{j=1}^J \bar{q}_j \lambda^{-1} \bar{q}_j  = \sum_{j=1}^J w_j
\end{align*}
Where \(w_j=\left.\bar{q}_j\right.^{-2}/\lambda_j\). Moreover \(\trans{\Vone} Q D^{-1} \trans{Q}\) is a vector with \(k\)-th elements \(\sum_{j=1}^J \bar{q}_j \lambda_j^{-1} q_{kj}=\sum_{j=1}^J w_j q^*_{kj}\) where \(q_{kj}^*=q_{kj}/\bar{q}_j\)

\bigskip

Therefore 
\begin{align*}
\PsiHat_{\text{GLS}}
= \frac{1}{\sum_{j=1}^J w_j} \sum_{j=1}^J w_j \trans{q_j^*} \VpsiHat
\end{align*}

\subsection{Via joint modeling}
\label{SM:GLSefficient}

Consider the simple case of a single continuous brain measurement (\(R=1\)), a single binary exposure (\(P=1\)) with equal probability of being \(0\) and \(1\), no covariate (\(C=0\)), and no missing value. We can use a joint linear model:
\begin{align*}
Y_{ij}=\alpha_j+\beta X_i + \varepsilon_{ij} \text{ where } \left(\varepsilon_{i1},\ldots,\varepsilon_{iJ}\right) \sim \Gaus\left(0,\Sigma_{\varepsilon}\right)
\end{align*}

Denote by \(\VmuHat_g=\left(\muHat_{g1},\ldots,\muHat_{gJ}\right)\) the vector empirical mean in each group (i.e. one relative to \(X=1\) another to \(X=0\)) of sample size \(\frac{n}{2}\). Since the mean and variance are sufficient statistics in Gaussian models, the joint linear model is equivalent to:
\begin{align*}
\muHat_{gj}=\alpha_j+\beta X_g + e_{gj} \text{ where } \left(e_{g1},\ldots,e_{gJ}\right) \sim \Gaus\left(0,2\Sigma_{\varepsilon}/n\right)
\end{align*}
Denote by \(\Delta \VmuHat=\left(\Delta \muHat_{1},\ldots,\Delta  \muHat_{J}\right)\) the vector of difference in mean, the previous model implies: 
\begin{align*}
\Delta \muHat_{j}=\beta + \epsilon_{j} \text{ where } \left(\epsilon_{1},\ldots,\epsilon_{J}\right) \sim \Gaus\left(0,4\Sigma_{\varepsilon}/n\right)
\end{align*}
whose maximum likelihood solution is \(\widehat{\beta}=\left(\trans{\Vone} \SigmaHat_{\epsilon}^{-1}\Vone\right)^{-1} \trans{\Vone} \SigmaHat_{\epsilon}^{-1} \Delta \VmuHat\). 

\bigskip

Now consider the GLS estimator of the common exposure effect \(\VpsiHat_{\text{GLS}}
= \left(\trans{\Vone} \SigmaHat_{\PsiHat}^{-1}\Vone\right)^{-1} \trans{\Vone} \SigmaHat_{\PsiHat}^{-1} \VpsiHat\) based on the pipeline specific Ordinary Least Squares (OLS) estimators of the exposure effect \(\VpsiHat=\left(\psiHat_1,\ldots,\psiHat_J\right)\). Denote by \(\Theta_j=\left(\alpha_j,\psi_j\right)\) the mean parameters of each pipeline specific model and \(\sigma^2_j\) the residual variance parameter. We have:
\begin{itemize}
    \item \(\VpsiHat = \left(\trans{X}X\right)^{-1}\trans{X} Y_{.j}\) where \(Y_{.j}\) denotes the brain measurements across individuals relative to the \(j\)-th pipeline.
    \item \(\SigmaHat_{\PsiHat}\) has elements \(\sum_{i=1}^n \trans{\left(Y_{ij} - X_i \Theta_j \right)}\left(Y_{ij} - X_i \Theta_j \right)\left(X_{i} \left(\trans{X}X\right)^{-1}\right)^2 \trans{c}\) where \(c=(0,1)\). This follows from the fact the the score relative to \(\psiHat_j\) is \(\frac{1}{\sigma_j^2} \trans{X}\left(Y_j - X \Theta_j\right)c\) and the variance covariance \(\sigma_j^2 \left(\trans{X}X\right)^{-1}\).
\end{itemize}
With a single binary covariate (and an intercept), \(\trans{X}X=\begin{bmatrix}
    n & n/2 \\ n/2 & n
\end{bmatrix}\) whose inverse is \(\begin{bmatrix}
    2/n & -2/n \\ -2/n & 4/n
\end{bmatrix}\)
Hence the second element of \(X_i(\trans{X}X)^{-1}\) is either \(-2/n\) or \(2/n\) so \(\left(X_i(\trans{X}X)^{-1}\right)\trans{c}=4/n\). Therefore, assuming that the observations are sorted by group:
\begin{align*}
    \psiHat_j &= \frac{n}{2} \sum_{i=1}^{n/2} Y_{ij} - \frac{n}{2} \sum_{i=n/2 +1}^{n} Y_{ij} = \Delta \muHat_{j} \\
    \SigmaHat_{\psiHat} &= \frac{4}{n^2} \sum_{i=1}^{n} \trans{\left(Y_{i.} - X \Theta\right)}\left(Y_{i.} - X \Theta\right)  = \frac{4}{n} \SigmaHat_{\varepsilon}\end{align*} So \(\VpsiHat_{\text{GLS}}=\left(\trans{\Vone} \SigmaHat_{\varepsilon}^{-1}\Vone\right)^{-1} \trans{\Vone} \SigmaHat_{\varepsilon}^{-1} \Delta \VmuHat\). If the common effect models holds (i.e. \(\psiHat_1=\ldots=\psiHat_J=\beta\)) then \(\SigmaHat_{\varepsilon}\) is an unbiased estimate of \(\Sigma_{\varepsilon}\) so \(\VpsiHat_{\text{GLS}}\) is asymptotically equivalent to the ML estimator \(\widehat{\beta}\) and thus asymptotically efficient.

\end{subappendices}

\end{document}